\documentclass[10pt,twocolumn,letterpaper]{article}

\usepackage[pagenumbers]{iccv} 

\definecolor{iccvblue}{rgb}{0.21,0.49,0.74}

\definecolor{gred}{RGB}{219,68,55}
\definecolor{gblue}{RGB}{66,133,244}

\usepackage[pagebackref,breaklinks,colorlinks,allcolors=iccvblue]{hyperref}
\usepackage{xcolor}         
\usepackage{caption}
\usepackage{amsmath}
\usepackage{multirow} 

\def\paperID{6814} 
\def\confName{ICCV}
\def\confYear{2025}

\title{Modification Takes Courage: \\ Seamless Image Stitching via Reference-Driven Inpainting}

\author{Ziqi Xie, Xiao Lai, Weidong Zhao, Siqi Jiang, Xianhui Liu, Wenlong Hou\\
Tongji University\\
}

\begin{document}
\twocolumn[{%
\renewcommand\twocolumn[1][]{#1}%
\maketitle
\begin{center}
\centering
\captionsetup{type=figure}
\includegraphics[width=1.0\textwidth]{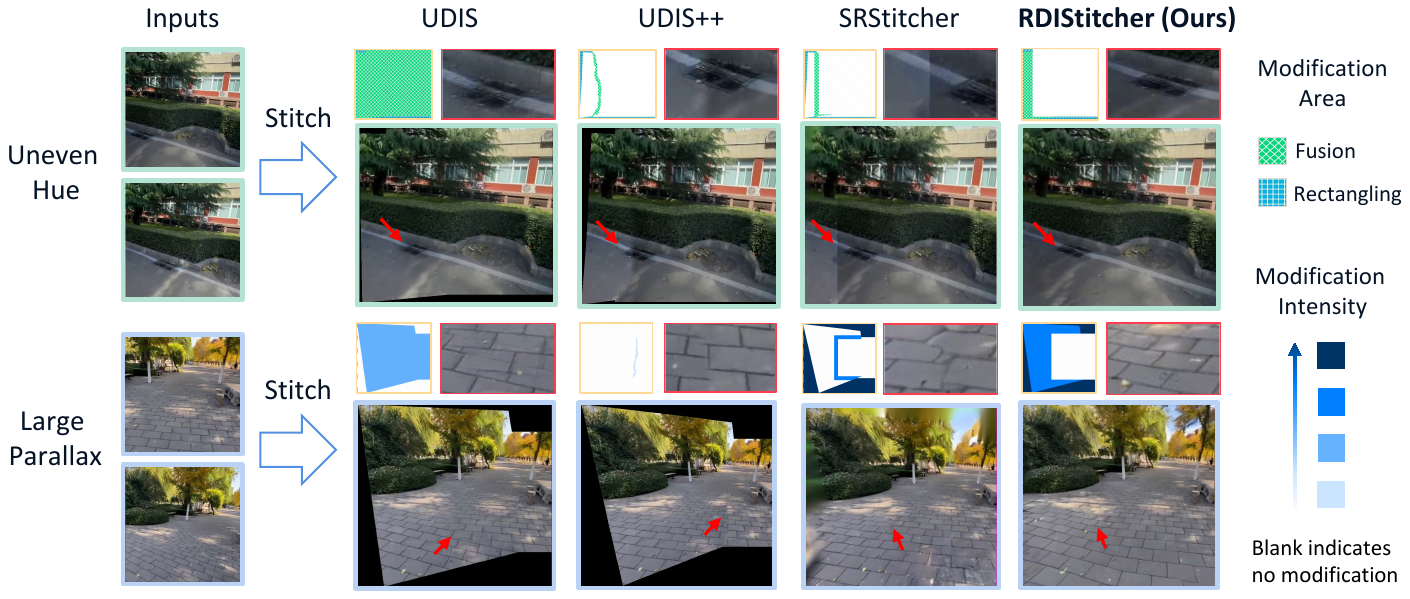}
\captionof{figure}{Different solutions to image fusion in image stitching. Our method reformulates the fusion and rectangling tasks as a reference-driven inpainting model. By boldly using a larger fusion modification area compared to UDIS++ \cite{nie2023parallax} and SRStitcher \cite{xie2024reconstructing}, and applying a stronger modification intensity than UDIS \cite{nie2021unsupervised} and UDIS++, we achieve a significant advancement in seamless image stitching, particularly in the challenging scenarios involving uneven hue and large parallax. Notice that UDIS and UDIS++ are three-stage architecture methods requiring additional rectangling models to complete the stitching process. Therefore, the rectangling areas for these methods are left blank in the figure.}
\label{fig:top}
\end{center}%
}]

\begin{abstract}
Current image stitching methods often produce noticeable seams in challenging scenarios such as uneven hue and large parallax. To tackle this problem, we propose the Reference-Driven Inpainting Stitcher (RDIStitcher), which reformulates the image fusion and rectangling as a reference-based inpainting model, incorporating a larger modification fusion area and stronger modification intensity than previous methods. Furthermore, we introduce a self-supervised model training method, which enables the implementation of RDIStitcher without requiring labeled data by fine-tuning a Text-to-Image (T2I) diffusion model. Recognizing difficulties in assessing the quality of stitched images, we present the Multimodal Large Language Models (MLLMs)-based metrics, offering a new perspective on evaluating stitched image quality. Compared to the state-of-the-art (SOTA) method, extensive experiments demonstrate that our method significantly enhances content coherence and seamless transitions in the stitched images. Especially in the zero-shot experiments, our method exhibits strong generalization capabilities. Code: \url{https://github.com/yayoyo66/RDIStitcher}
\end{abstract}    
\section{Introduction}
\label{sec:intro}
Image stitching is a fundamental problem in computer vision, which aims to seamlessly integrate multiple images captured from different perspectives into a wide field-of-view composite image \cite{lin2011smoothly,zhang2014parallax,gao2011constructing}. Image fusion is the core stage of the image stitching pipeline and focuses on combining aligned images without visible seams or artifacts. However, as illustrated in \textcolor{red}{Fig.}\ref{fig:top}, this stage encounters two major challenges: (1) \textbf{Uneven hue.} Due to variations in atmospheric lighting conditions and camera settings, images taken from different viewpoints of the same scene may display inconsistent hues. When fusing images with significant hue differences, visible seams will likely appear in the stitched image. (2) \textbf{Large parallax.} Large parallax refers to the significant difference in the relative positions of objects in a scene when captured from different viewpoints. Existing homography-based registration methods \cite{nie2020view,nguyen2018unsupervised,li2017quasi} struggle to accurately align images in large parallax scenes, leading to noticeable artifacts and misalignment of content in the stitched images.

\begin{figure}[t]
  \centering
   \includegraphics[width=0.96\linewidth]{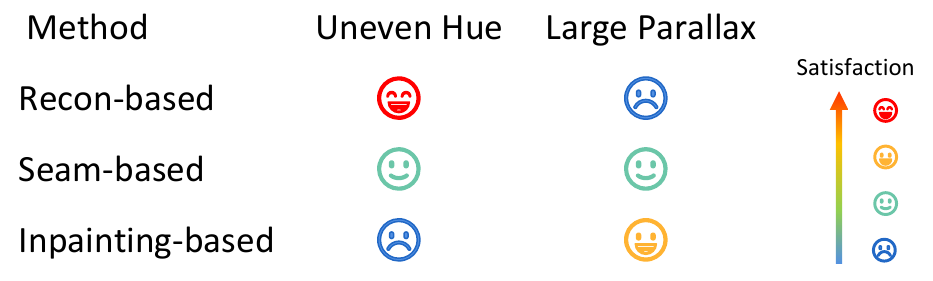}
   \caption{A user experience survey of the recon-based method UDIS \cite{nie2021unsupervised}, the seam-based method UDIS++ \cite{nie2023parallax}, and the inpainting-based method SRStitcher \cite{xie2024reconstructing} on uneven hue and large parallax scenes. Please see the Supplementary Material for more details.}
   \label{fig:suvey}
\end{figure}

Current image fusion methods can be divided into three categories, including reconstruction-based (recon-based), seam-based, and inpainting-based. Recon-based methods \cite{nie2020view,nie2022learning,nie2021unsupervised} use pixel-by-pixel reconstruction to smooth the fused image, effectively handling scenes with uneven hues. However, in large parallax scenarios, recon-based methods can introduce notable artifacts, which degrade image quality. Additionally, seam-based methods \cite{nie2023parallax,cheng2023deep,li2022automatic} work by identifying optimal seams for image fusion. Nevertheless, these methods heavily rely on the assumption that perfect seams exist, which often fails to hold true in uneven hue and large parallax scenarios. Finally, the inpainting-based method \cite{xie2024reconstructing} proposes to modify the fusion area to improve fusion effects. Unfortunately, the existing method is conservative in the size selection of modification fusion areas, so it is difficult to deal with uneven hue and large parallax scenarios. To detail the advantages and disadvantages of the three methods in different challenge scenarios, we conducted a small user experience survey presented in the \textcolor{red}{Fig.}\ref{fig:suvey}. 

To address the limitations of current methods, we propose a key principle: \textbf{\textit{Modification takes courage, including area size and intensity.}} We develop the RDIStitcher, which utilizes a larger modified area for fusion than previous seam-based and inpainting-based methods. Compared to recon-based methods, RDIStitcher applies stronger modification intensity.  

However, larger and stronger modifications come at a cost, which is introducing more content instability into the stitched image. Due to the shortage of labeled data, SRStitcher is unable to train the model and can only maintain content consistency before and after stitching by limiting the size of the fusion modification area. Therefore, to implement RDIStitcher, we propose a self-supervised training method. Specifically, we leverage pre-knowledge from an unlabeled image stitching dataset \cite{nie2021unsupervised} by using a pre-trained registration model to generate pseudo-stitching images based on single-view images. Subsequently, we apply a large-scale pre-trained T2I diffusion model \cite{stablediffusion2inpainting} to learn the restoration of the single-view image based on the pseudo-stitching images, effectively teaching the model the new concept of stitching. Our experiments demonstrate that this self-supervised training method achieves remarkable results with high generalization ability, requiring only a few parameters to fine-tune the T2I model.

After designing the model and training method, the final task is to measure the stitched image quality in challenging scenarios without ground truth. The previous works UDIS \cite{nie2021unsupervised} and UDIS++ \cite{nie2023parallax} rely solely on small-scale user evaluations to assess stitched image quality, which is costly and lacks comprehensiveness. In addition, SRStitcher \cite{xie2024reconstructing} introduces the No-Reference Image Quality Assessment (NR-IQA) metrics. However, existing NR-IQA techniques \cite{su2020blindly,wang2023exploring} have significant flaws when applied to assessing stitched image quality, especially in detecting fine-grained stitching issues \cite{xie2024reconstructing}. To address these challenges, we develop assessment methods based on MLLMs for stitched images, including the Single-Image Quality Score (SIQS) and the Multi-Image Comparative Quality Score (MICQS). 

We summarize the main contributions as follows:

\begin{itemize}
 \item \textit{We reformulate the fusion and rectangling tasks as a reference-driven inpainting model.} This model achieves remarkable stitching effects in challenging scenarios including uneven hues and significant parallax, while preserving the original structure and content of the input images. (\textcolor{red}{Sec.}\ref{sec:Definition} and \textcolor{red}{Sec.}\ref{sec:Framework})
 \item \textit{We introduce a self-supervised training method that enables RDIStitcher to be trained without the need for labeled data.} This method requires fine-tuning only a small number of parameters in a large-scale pre-trained T2I model, resulting in low hardware requirements. To our knowledge, this is the first unsupervised training method for the rectangling problem. (\textcolor{red}{Sec.}\ref{sec:STM})
 \item \textit{We propose quality metrics for assessing stitched images by MLLMs.} By incorporating the MLLMs into the image stitching domain for the first time, we offer a pioneering research perspective for the evaluation of stitched images. (\textcolor{red}{Sec.}\ref{sec:LLMmetrics})
\end{itemize}


\section{Related Work}
\label{sec:relatedwork}

\subsection{Image Stitching} \label{sec:relatedworkIS}
The previous learning-based image stitching methods typically follow a three-stage architecture: (1) Registration stage:  In this stage, works \cite{nie2020view,nguyen2018unsupervised,li2017quasi} estimate the homography between the input images using networks. Then, they warp the input images to a unified coordinate system to obtain aligned images. (2) Fusion stage: This stage focuses on fusing the aligned images into a seamless composite image. Existing methods can be divided into recon-based methods \cite{nie2020view,nie2022learning,nie2021unsupervised} and seam-based methods \cite{nie2023parallax,cheng2023deep,li2022automatic}. (3) Rectangling stage: The objective of this stage is to transform irregularly edged stitched images into regular rectangular formats. Current rectangling methods \cite{nie2022deep,wang2024sdr,xie2024rectanglinggan,zhou2024recdiffusion} are all supervised and rely on labeled datasets.

Beyond the three-stage architecture, the inpainting-based method SRStitcher \cite{xie2024reconstructing} simplifies the image stitching pipeline by integrating fusion and rectangling into a unified model. SRStitcher is implemented based on a large-scale pre-trained generative model, requiring no additional training or fine-tuning. However, this methodology implies that the performance of SRStitcher is entirely dependent on the capabilities of the pre-trained model. Consequently, SRStitcher adopts a highly conservative strategy in managing the fusion region size to avoid excessive alterations, but this results in noticeable seams in challenging scenarios. Furthermore, the generalization capabilities of SRStitcher are inherently limited by the pre-trained model, making it difficult to extend to domains such as medical imaging \cite{bergen2014stitching,adwan2016new} and remote sensing \cite{wang2024spatial,wan2020drone}.

Our method builds on the framework established by SRStitcher and further advances it by proposing a reference-driven inpainting model for fusion and rectangling.  Additionally, we implement a self-supervised training method that is specifically designed for this new modeling paradigm. This enhancement increases the scalability and adaptability of the inpainting-based image stitching pipeline. Furthermore, our method represents a significant advancement in addressing the rectangling problem, as it offers the first training method that does not require labeled data.

\subsection{Reference-Driven Inpainting}

Image inpainting aims to fill missing image regions with coherent results \cite{dong2022incremental,zeng2020high,li2022mat}. Reference-driven inpainting \cite{yang2023paint,zhou2021transfill,tang2024realfill,cao2024leftrefill} is a new subfield of image inpainting research that focuses on filling missing regions leveraging content from reference images. Although some reference-driven inpainting methods \cite{zhou2021transfill,tang2024realfill} show potential for applications in image stitching, they have not developed a complete training pipeline or conducted comprehensive experimental validation.

In contrast, our method introduces the first image stitching pipeline based on reference-driven inpainting, encompassing model design, training, and validation, as well as introducing the novel evaluation metrics. Our method employs the input and output construction strategy of LeftRefill \cite{cao2024leftrefill}, but differs significantly in terms of the training method, model structure, and the number of parameters.

\section{Method}
\subsection{Problem Definition} \label{sec:Definition}

\begin{figure*}[ht]
\centering
\includegraphics[width=0.96\linewidth]{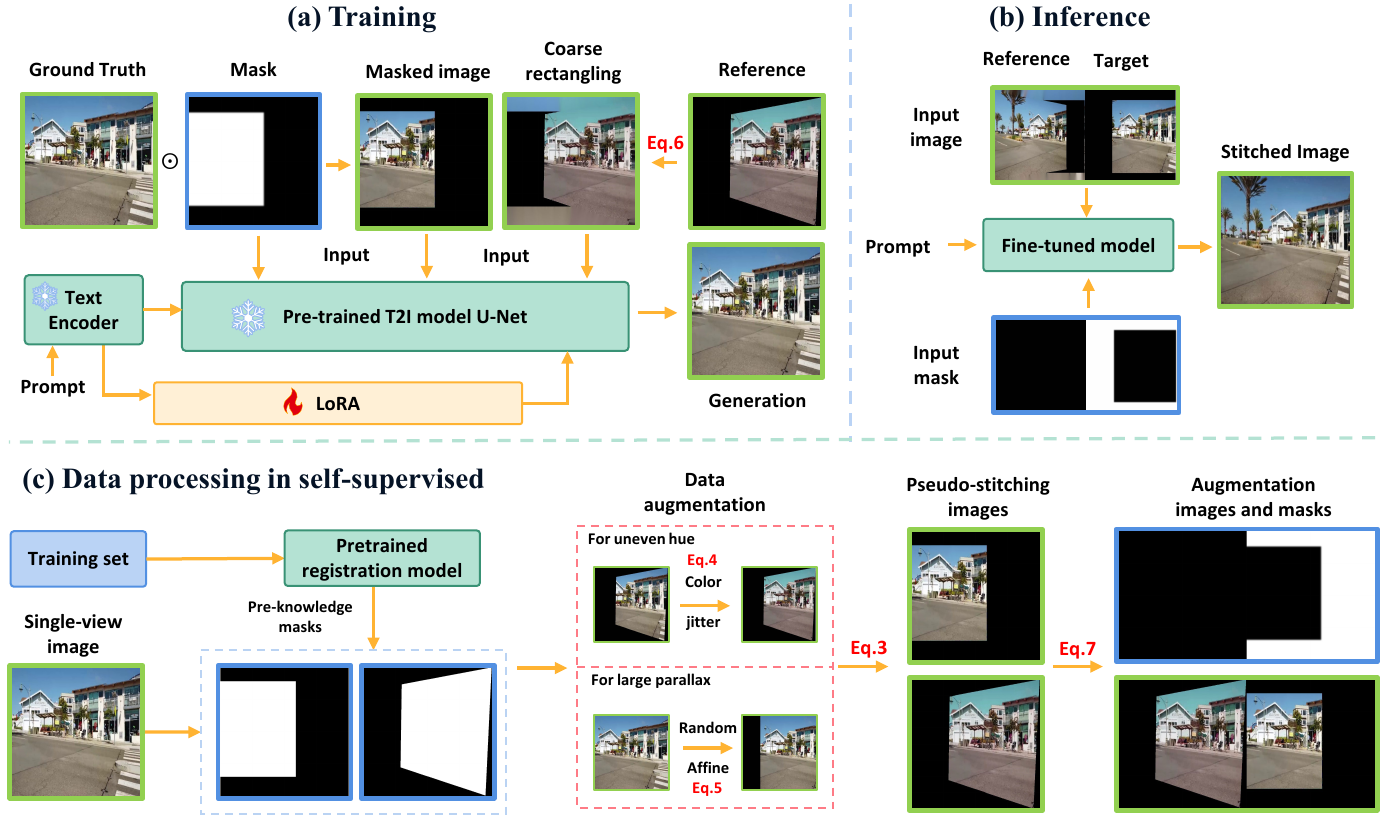}
\caption{The framework of RDIStitcher. (a) Training. For the sake of clarity in the presentation, the input images and masks are simplified. (b) Inference. Details on the specific input structure. (c) Data processing in self-supervised. Details on the self-supervised training method.}
\label{fig:framework}
\end{figure*}
        
Given a pair of input images: a reference image and a target image, defined as $I_{r}, I_{t} \in \mathbb{R}^{H \times W}$, where $H$ and $W$ are the height and width, respectively. Similar to SRStitcher \cite{xie2024reconstructing}, the image stitching pipeline designed in this paper is a two-stage architecture. 

In the first stage, the homography matrix $\mathcal{H}$ between the input images is estimated by a registration model. Then, the $I_{r}$ is aligned with the $I_{t}$ by a warp function $\mathcal{W}(\cdot)$ based on $\mathcal{H}$. The process of obtaining the aligned images $I_{wr}$ and $I_{wt}$ can be expressed as \textcolor{red}{Eq.}\ref{eq:align}.

\begin{equation}
I_{wr},I_{wt} = \mathcal{W}(I_{r},\mathcal{H}),\mathcal{W}(I_{t},\mathbb{I}),
\label{eq:align}
\end{equation}

where, $\mathbb{I}$ is the identity matrix. The masks $M_{wr}$ and $M_{wt}$ corresponding to the aligned images can also be obtained by the \textcolor{red}{Eq.}\ref{eq:align}, except that the input images is replaced by the all-ones matrixes of the same size.

In the second stage, we reformulate the fusion and rectangling problems as a reference-driven inpainting model. Given a target image $I_{wt}$ and its corresponding mask $M_{wt}$, the objective is to train a model $\epsilon_{\theta}(\cdot)$ that inpaints the image $I_{wt}$ in accordance with the mask $\overline{M}_{wt} = 1-M_{wt}$ to obtain the stitched image $I_{s}$, where the inpainting process is constrained by the content of the reference image $I_{wr}$, as represented in \textcolor{red}{Eq.}\ref{eq:rdimodel}.

\begin{equation}
I_{s} = I_{wt}\odot M_{wt} + \epsilon_{\theta}(I_{wr},I_{wt})\odot \overline{M}_{wt},
\label{eq:rdimodel}
\end{equation}

where, $\odot$ denotes pixel-wise multiplication. Different from SRStitcher \cite{xie2024reconstructing}, RDIStitcher employs the $\mathcal{W}(\cdot)$ from UDIS \cite{nie2021unsupervised} instead of UDIS++ \cite{nie2023parallax}. Because the $\mathcal{W}(\cdot)$ of UDIS++ occasionally introduces huge local distortions into the aligned images, which could potentially disrupt the fitting process that relies on undistorted images. In addition, we select the image warped by $\mathbb{I}$ as the target image, it is also a consideration of the degree of image distortion.

\subsection{Framework of RDIStitcher} \label{sec:Framework}
We propose the introduction of a large-scale pre-trained T2I model to enhance the generalization capabilities in the design of $\epsilon_{\theta}(\cdot)$. Furthermore, we develop a training method based on self-supervised paradigms to improve the practical feasibility. In general, the framework of RDIStitcher is illustrated in \textcolor{red}{Fig.}\ref{fig:framework}. 

Similar to the input structure of the T2I model \cite{stablediffusion2inpainting}, the input of RDIStitcher consists of three parts: image $\mathbf{I}$, mask $\mathbf{M}$ and masked image $\mathbf{I_{m}}= \mathbf{I} \odot (1-\mathbf{M})$. Specifically, the image $I$ is a composite consisting of the aligned target image $I_{wt}$ and the coarse rectangling aligned reference image $I^{cr}_{wr}$, that is $\mathbf{I}=\mathsf{Concat}(I^{cr}_{wr},I_{wt})$. Accordingly, the mask $\mathbf{M}$ consists of an all-zero matrix $M_0$ and the gradient aligned target mask $\mathfrak{M}_{wt}$, that is $\mathbf{M}=\mathsf{Concat}(M_0,\mathfrak{M}_{wt})$ (The coarse rectangling and gradient mask are detailed in \textcolor{red}{Sec.}\ref{sec:STM}). Conversely, the output size of the model is equivalent to the input size, and the left keeps $I^{cr}_{wr}$, but the right is generated to $I_{s}$. 

To enable the T2I model to perform specific image stitching tasks, we incorporate the unique identifier from DreamBooth \cite{ruiz2023dreambooth} into our method. Specifically, we define a special text prompt, denoted as $\mathcal{P}$, which guides the model in carrying out reference-driven inpainting operations. During training, we use the Low-Rank Adaptation (LoRA) \cite{hu2022lora} to fine-tune both the text encoder and the U-Net of the T2I model, effectively reducing memory usage.

\subsection{Self-Supervised Training Method} \label{sec:STM}

A significant challenge in training $\epsilon_{\theta}(\cdot)$ is the extreme scarcity of available data, particularly labeled datasets for image stitching \cite{nie2021unsupervised}. The lack of labeled data makes it difficult to directly apply mainstream supervised learning paradigms. To overcome this challenge, we turn to the self-supervised learning paradigm, which aims to generate meaningful training signals by exploiting the inherent structure or relationships present in the input data, without the need for explicit annotations \cite{he2022masked,noroozi2016unsupervised}. 

To learn image stitching concepts, we hypothesize: \textbf{\textit{We can use a single-view image to simulate the pseudo-stitched images and then train a model to restore the original image based on the pseudo-stitched images. Through this process, the model can establish an understanding of the underlying stitching relationships in the data. Subsequently, we can leverage the strong generalization capabilities of the large-scale T2I model to generalize this knowledge to real stitching data.}}

As shown in \textcolor{red}{Fig.}\ref{fig:framework} (c), we collect the stitching mask distribution from a large-scale image stitching dataset UDIS-D \cite{nie2021unsupervised} by a pre-trained registration model \cite{nie2021unsupervised}.The pre-knowledge mask distribution is defined as $\mathsf{M}_N =\{(M^1_{wr},M^1_{wt}),(M^2_{wr},M^2_{wt}),...,(M^N_{wr},M^N_{wt})\}$, where $N$ is the number of samples in UDIS-D training set $D_{train}$. Then, suppose $I_{sg}$ is a single-view image in the $D_{train}$ and $(M^i_{wr},M^i_{wt}), i \in [1,N]$ is a random mask set from $\mathsf{M}_N$, the pseudo-stitching reference and target images $\tilde{I}_{wr},\tilde{I}_{wt}$ can be obtained by \textcolor{red}{Eq.}\ref{eq:pseudo}.

\begin{equation}
\tilde{I}_{wr},\tilde{I}_{wt} = I_{sg} \odot M^i_{wr}, I_{sg} \odot M^i_{wt}.
\label{eq:pseudo}
\end{equation}

However, the pseudo-stitching images produced by the above method are inherently \textit{perfectly correct} and lack challenging scenarios. To improve the robustness, we design two data augmentation strategies to simulate color differences of uneven hue conditions and misalignments of large parallax conditions.

\noindent
\textbf{Data augmentation for uneven hue.} To simulate color differences, we add random color jitter $\mathsf{ColorJitter}(\cdot)$ \cite{paszke2019pytorch} with probability $p_{cj}$ to the pseudo-stitching reference image $\tilde{I}_{wr}$, as formulated in \textcolor{red}{Eq.}\ref{eq:colorjitter}.

\begin{align}
\tilde{I}_{wr} =\left\{ \begin{array}{l}
\mathsf{ColorJitter}(\tilde{I}_{wr},\mathbf{E}_{cj}) \quad \text{with } p_{cj},\\
\tilde{I}_{wr} \qquad\qquad\qquad\quad \text{with } 1 - p_{cj},\\
\end{array} \right.
\label{eq:colorjitter} 
\end{align}

where, $\mathbf{E}_{cj}=\{e_b, e_c,e_s,e_h\}$ are the hyper-parameters of the brightness, contrast, saturation and hue adjustments.

\noindent
\textbf{Data augmentation for large parallax.} The registration errors induced by large parallax are manifested differently across various warping methods. In UDIS++ \cite{nie2023parallax}, these errors result in pronounced local distortions, while in UDIS \cite{nie2021unsupervised}, they lead to content misalignment near the seams. Obviously, content misalignment is comparatively simpler to simulate, so we select the $\mathcal{W}(\cdot)$ of UDIS as the foundation for designing RDIStitcher. To simulate misalignments, a random affine transformation $\mathsf{Aff}(\cdot)$ with probability $p_{at}$ is applied to the $\tilde{I}_{wr}$ and its corresponding mask $M^i_{wr}$, as demonstrated in \textcolor{red}{Eq.}\ref{eq:affine}.

\begin{align}
\tilde{I}_{wr} =\left\{ \begin{array}{l}
\mathsf{Aff}(I_{sg},\mathcal{M}(t_x,t_y)) \odot M^i_{wr}\quad \text{with } p_{at},\\
I_{sg}  \odot M^i_{wr} \qquad\qquad\qquad\quad \text{with } 1 - p_{at},\\
\end{array} \right.
\label{eq:affine} 
\end{align}

where, $\mathcal{M}(\cdot)$ is a $2\times 3$ affine transformation matrix whose horizontal $t_x \in [-(W^*-x_{max}),x_{min}]$ and vertical $t_y \in [-(H^*-y_{max}),y_{min}]$ translations are random values, $W^*$ and $H^*$ are the width and height of $I_{sg}$. Also, $x_{max}$, $x_{min}$, $y_{max}$ and $y_{min}$ are coordinates of the smallest enclosing rectangle of $M^i_{wr}$.

\noindent
\textbf{Coarse rectangling and gradient mask.} The coarse rectangling and gradient mask are originally introduced by SRStitcher \cite{xie2024reconstructing}. Coarse rectangling aims to reduce the likelihood of generating abnormal content in the rectangling regions by incorporating weak priors. And, the gradient mask is utilized to smooth the seam regions. However, SRStitcher claims that coarse rectangling introduces a local blurring side effect that degrades image quality. To address this issue, we propose a self-supervised training method that allows the model to establish the relationship between the weak priors and the concrete images, effectively mitigating the local blur problem. Additionally, we reconstruct the gradient mask based on the RDIStitcher input mask to enhance image cohesion in the seam regions. Specifically, \textcolor{red}{Eq.}\ref{eq:cr} details the process of obtaining the coarse rectangling image $\tilde{I}^{cr}_{wr}$ by the alexandru telea algorithm $\mathsf{Telea}(\cdot)$ \cite{telea2004image}.

\begin{equation}
\tilde{I}^{cr}_{wr} = \mathsf{Telea}(\tilde{I}_{CF},M^i_{wr}\vee M^i_{wt},R),
\label{eq:cr}
\end{equation}

where, $\tilde{I}_{CF}=\tilde{I}_{wr}+\tilde{I}_{wt}\odot (1-(M^i_{wr} \& M^i_{wt}))$, $\vee$ and $\&$ denote the bitwise OR and AND operators, and $R$ is the hyper-parameter to control the radius. Furthermore, We use \textcolor{red}{Eq.}\ref{eq:gm} to generate the seam gradient target mask $\mathfrak{M} ^{i}_{wt}$.

\begin{equation}
\mathfrak{M} ^{i}_{wt} = \mathsf{Blur}(\mathsf{Dilation}(M^i_{wt},K_d),K_g),
\label{eq:gm}
\end{equation}

where, $\mathsf{Dilation}(\cdot)$ is the dilation operation \cite{2015opencv} with kernel $K_d$, and $\mathsf{Blur}(\cdot)$ is the Gaussian blur operation \cite{2015opencv}  with kernel $K_g$.

\noindent
\textbf{Objective function.} RDIStitcher is based on Denoising Diffusion Probabilistic Model (DDPM) \cite{ho2020denoising}, where the $\epsilon_{\theta}(\cdot)$ is transformed into a noisy prediction model. Suppose the latents $\mathbf{x}_0,\mathbf{x}_1,...,\mathbf{x}_T$ are derived by adding Gaussian noise to the original data $\mathbf{x}_0 \sim q(\mathbf{x}_0)$ with $T$ steps, where $\mathbf{x}_0=\mathsf{ImageEncoder}(\mathsf{Concat}(\tilde{I}^{cr}_{wr},I_{sg}))$. The objective function $\mathcal{L}$ of $\epsilon_{\theta}(\cdot)$ in RDIStitcher is defined as in \textcolor{red}{Eq.}\ref{eq:obj}.

\begin{equation}
\mathcal{L} = \mathbb{E}_{\vartheta} \left [  \left \| \epsilon- \epsilon_\theta (\mathbf{x}_t,t,\mathcal{P},\mathbf{M}^i,\overline{\mathbf{M}^i} \odot \mathbf{x}_0) \right \|^2 \right ], 
\label{eq:obj}
\end{equation}

where, $\mathbf{M}^i=\mathsf{Concat}(M_0,\mathfrak{M} ^{i}_{wt})$, $\overline{\mathbf{M}^i}=1-\mathbf{M}^i$, $t \in [1,T]$, $\vartheta = \{\mathbf{x}_0,t,\epsilon,\mathbf{M}^i\}$, and $\epsilon \sim \mathcal{N}(0,1)$.
\section{Experiment}
\subsection{Experimental Setup}

\textbf{Dataset.} 
Most of the experiments in this section are conducted on the unsupervised dataset UDIS-D \cite{nie2021unsupervised}. To the best of our knowledge, UDIS-D is the only publicly available dataset in learning-based image stitching. To provide a more comprehensive evaluation, we conduct cross-dataset experiments in a zero-shot manner on three traditional datasets, including the APAPdataset \cite{zaragoza2013projective}, the REWdataset \cite{li2017parallax}, and the SPWdataset \cite{liao2019single}. However, since these traditional datasets are small and contain only tens of samples, performing quantitative evaluations on them would not produce meaningful results. Therefore, this work focuses on a qualitative evaluation to demonstrate the differences in the generalization capabilities on traditional datasets.

\noindent
\textbf{Baseline method.} 
Selecting appropriate baseline methods is a little difficult, because our method is the first to propose a fusion and rectangling unified model that can be trained on the unsupervised image stitching dataset, a groundbreaking advancement in the image stitching. As mentioned in \textcolor{red}{Sec.}\ref{sec:relatedworkIS}, UDIS \cite{nie2021unsupervised} and UDIS++ \cite{nie2023parallax} can only perform up to the fusion stage and require integration with another rectangling model to complete the entire image stitching pipeline. However, the SOTA rectangling models \cite{nie2022deep,wang2024sdr,xie2024rectanglinggan,zhou2024recdiffusion} are all supervised and can only be trained on a task-specific dataset DIR-D \cite{nie2022deep}, rather than the real-world image stitching dataset UDIS-D. This limitation makes it difficult to compare these methods with RDIStitcher under relatively fair conditions. To address this issue, we propose a variant of RDIStitcher, called RDIStitcher-R, which works only as a rectangling model trained with the same settings as RDIStitcher. With this, we can combine the RDIStitcher-R with UDIS and UDIS++ to establish baselines. In summary, the baselines include the SOTA recon-based method UDIS with RDIStitcher-R (UDIS+R) and the SOTA seam-based method UDIS++ with RDIStitcher-R (UDISplus+R). The test results of RDIStitcher-R on DIR-D are SSIM 0.763 and PSNR 23.56, positioning its performance between the DeepRectangling \cite{nie2022deep} and Recdiffusion \cite{zhou2024recdiffusion}. Implementation details and validation experiments for RDIStitcher-R can be found in the Supplementary Material.

In addition, the baselines also include the SOTA inpainting-based method SRStitcher \cite{xie2024reconstructing}. Although the SRStitcher method does not involve any training or fine-tuning, which makes the comparison somewhat less fair, we believe it is necessary to include the results of this most important related work. And, we also implement our self-supervised training method on LeftRefill \cite{cao2024leftrefill} as a baseline.

\noindent
\textbf{Implement detail.} 
RDIStitcher is fine-tuned on a pre-trained T2I model \cite{stablediffusion2inpainting} with a batch size of 4, input image size 1024$\times$512, LoRA rank 8, learning rate 2e-4 for the U-Net, learning rate 4e-5 for the text encoder, and 10K iterations on two NVIDIA 4090 GPUs. And, all results of RDIStitcher described in the experiment are based on the pre-aligned images made by UDIS \cite{nie2021unsupervised}. For hyper-parameters, the $\mathbf{E}_{cj}$ and $p_{cj}$ of \textcolor{red}{Eq.}\ref{eq:colorjitter} are set to $\{0.2,0.2,0.2,0.1\}$ and 0.25; The $p_{at}$ of \textcolor{red}{Eq.}\ref{eq:affine} is set to 0.25; The $R$ of \textcolor{red}{Eq.}\ref{eq:cr} is set to 3; The $K_d$ and $K_g$ of \textcolor{red}{Eq.}\ref{eq:gm} are set to 10 and 15. For inference, the inference step is set to 50. More details can be found in Supplementary Material.

\subsection{MLLMs-based Metric} \label{sec:LLMmetrics}

\renewcommand\arraystretch{1.0}
\begin{table}[t]
\centering
\caption{An example of SIQS evaluated by Qwen-VL-Max.}
\label{tab:siqs}
\begin{tabular}{p{3in}}
\toprule
\multicolumn{1}{c}{
    \includegraphics[width=0.25\linewidth]{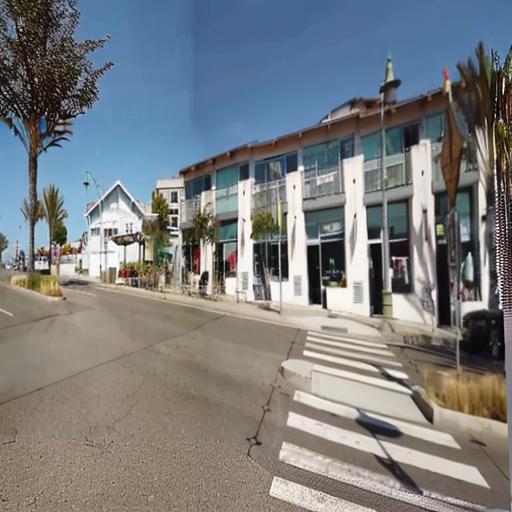}
} \\
\midrule
\textbf{Evaluation Prompt} \\

I need you to become a stitched image quality assessment evaluator. The evaluation process should be as objective and impartial as possible, giving specific ratings and reasons, including seam, brightness transition, distortion, clear and abnormal content, each aspect 2 points. \\
\textit{[The full example is in the Supplementary Material.]} \\
\midrule
\textbf{Reason} \\
- \textbf{Seam} The image shows slightly visible boundaries, particularly noticeable in the sky and along the edges of the buildings. (\textbf{\textcolor{gred}{1 points}}). \\
- \textbf{Brightness transition} The brightness transition in the image is relatively smooth. (\textbf{\textcolor{gred}{2 points}}). \\
- \textbf{Distortion} There are no noticeable distortions in the image. The straight lines of the buildings and the street appear to be accurate. (\textbf{\textcolor{gred}{2 points}}). \\
- \textbf{Clear} The buildings, palm trees, and streets are all distinguishable, and the details are visible. (\textbf{\textcolor{gred}{2 points}}). \\
- \textbf{Abnormal content} The image contains some unnatural elements, such as the seams. (\textbf{\textcolor{gred}{1 points}}). \\
\midrule
\textbf{Score} \\
The overall impression is that the image is a stitched panorama with noticeable flaws. (\textbf{\textcolor{gblue}{8 points}}).\\
\bottomrule
\end{tabular}
\end{table}

Assessing the quality of stitched images on unlabeled datasets has long been a challenging problem in the field. Previous works UDIS \cite{nie2021unsupervised} and UDIS++ \cite{nie2023parallax} rely solely on small-scale user evaluations to measure stitched image quality, which lacks comprehensiveness. Additionally, SRStitcher \cite{xie2024reconstructing} proves that the mainstream NR-IQA metrics \cite{su2020blindly,wang2023exploring} struggle to accurately evaluate stitched image quality, particularly in evaluating fine-grained stitching issues such as uneven hue and large parallax.

Quality measurement challenges are not unique to image stitching. After surveying the broader research landscape, we observe that many emerging domains, such as personalization generation \cite{avrahami2023break,zhang2023adding}, face similar measurement difficulties. In response, some studies \cite{wu2024comprehensive,zhang2023gpt,zhong2024multi} have proposed leveraging MLLMs to evaluate the quality of unlabeled images, leading to significant advancements in the field.

Motivated by these developments, we ask: \textbf{\textit{Can MLLMs be leveraged as effective evaluators to measure the quality of stitched images?}} To explore this, we propose the MLLMs-based stitched image evaluation metrics, including the Single-Image Quality Score (SIQS) and the Multi-Image Comparative Quality Score (MICQS).

To clarify the metrics, an example of SIQS is presented in \textcolor{red}{Table.}\ref{tab:siqs}, conducted by Qwen-VL-Max \cite{Qwen-VL}, denoted as SIQS-Q. In addition,  \textcolor{red}{Table.}\ref{tab:micqs} provides the MICQS evaluation comparing two stitched images, also performed by Qwen-VL-Max, denoted as MICQS-Q. To mitigate potential biases associated with relying on a single model, we introduce the model GLM-4V-Plus \cite{glm2024chatglm} as an additional evaluator (SIQS-G, MICQS-G). \footnote{We do not employ widely used MLLMs such as GPT-4V or Claude-3-Haiku, due to the limited access.}.
 
\renewcommand\arraystretch{1.0}
\begin{table}[t]
\centering
\caption{An example of MICQS evaluated by Qwen-VL-Max.}
\label{tab:micqs}
\begin{tabular}{p{3in}}
\toprule
\multicolumn{1}{c}{
    \includegraphics[width=0.25\linewidth]{figs/SIQStable.jpg}
    \hspace{25pt}
    \includegraphics[width=0.25\linewidth]{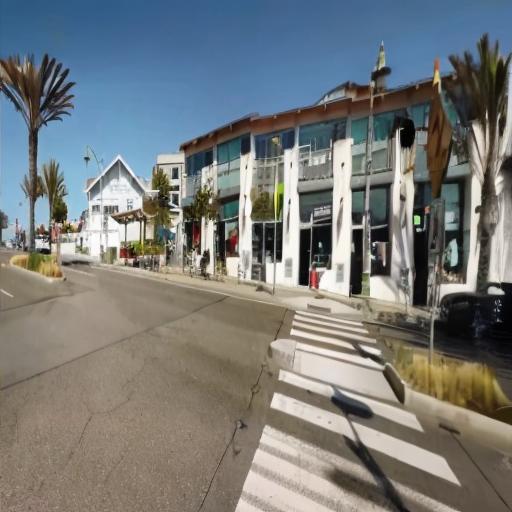}
} \\
\midrule
\textbf{Evaluation Prompt} \\
I need you to become a stitched image quality assessment evaluator. Compare the two input stitched images and choose which one you think is better. There are two choices: image 1 (left) is better, or image 2 (right) is better. \\
\textit{[The full example is in the Supplementary Material.]} \\
\midrule
\textbf{Reason} \\
-  \textbf{Seam} Image 2 has a smoother transition between the stitched sections compared to image 1, where the seam is more noticeable. \\
- \textbf{Brightness transition} Image 2 has a more consistent brightness level across the entire image, while image 1 shows a noticeable difference in brightness.\\
- ......\\
\midrule
\textbf{Conclusion} \\
\textbf{\textcolor{gblue}{Image 2 (right)}} is better.\\
\bottomrule
\end{tabular}
\end{table}

\subsection{Quantitative Evaluation}

In addition to the MLLMs-based image quality metrics, we also introduce the Content Consistency Score (CCS) from SRStitcher \cite{xie2024reconstructing} and the real user evaluation from UDIS \cite{nie2021unsupervised} with the same settings in quantitative evaluations.

\noindent
\textbf{Single-image evaluation.} \textcolor{red}{Table.}\ref{tab:quantitative} presents the comparative experimental results of SIQS-Q$\uparrow$, SIQS-G$\uparrow$, and CCS$\uparrow$ on the UDIS-D test set with 1,106 sample pairs. Our method surpasses other methods in terms of both quality and content consistency.

\begin{figure}[t]
\centering
\includegraphics[width=0.96\linewidth]{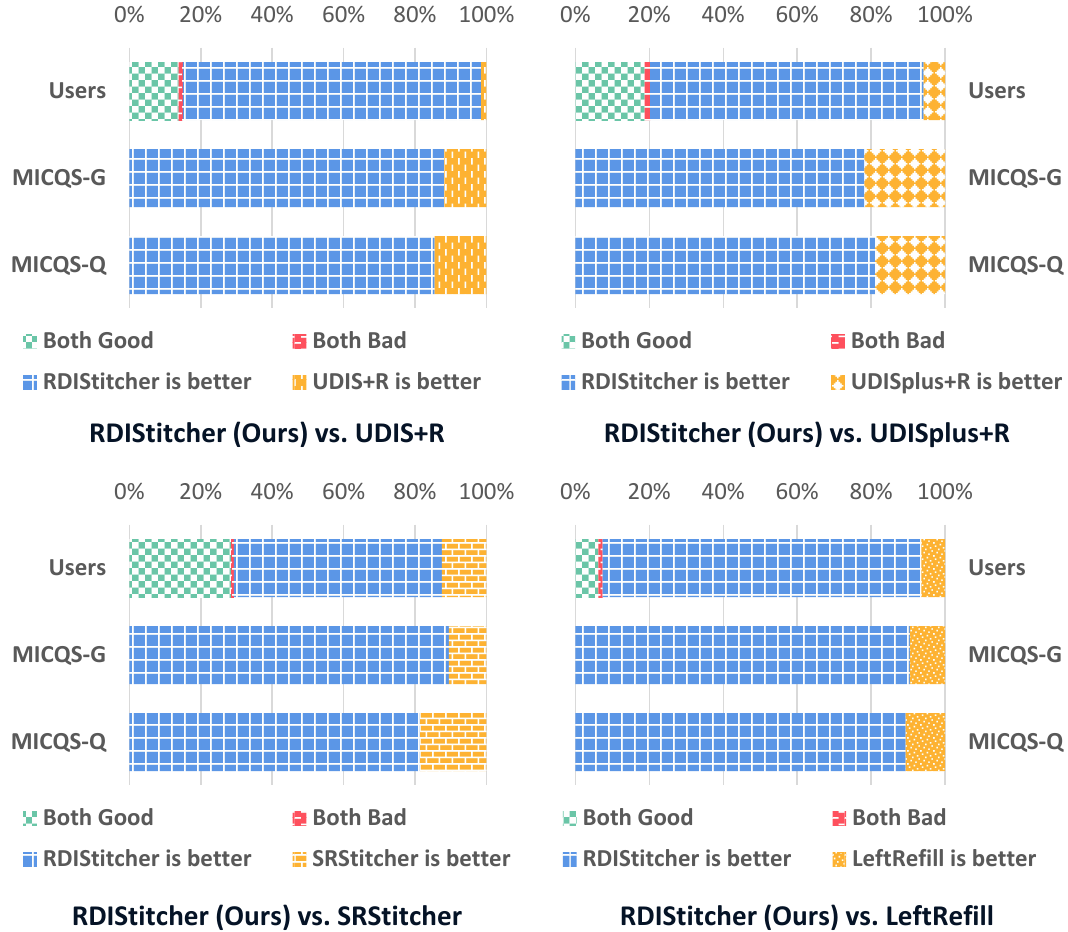}
\caption{Multi-image comparative evaluation results. We decide not to give MLLMs-based evaluators the \textit{both} option as we discover that they consistently favor \textit{both good}.}
\label{fig:Comparativequality}
\end{figure}

\noindent
\textbf{Multi-image comparative evaluation.} This method evaluates the stitched image quality in a contrastive manner, including MLLMs-based evaluators and real user evaluators. The results are presented in the \textcolor{red}{Fig.}\ref{fig:Comparativequality}, which illustrates that our method achieves significantly higher votes, clearly outperforming the methods.

\noindent
\textbf{Ablation study.} Our method is designed to be inherently interpretable at each module, which makes ablation experiments less critical. Therefore, ablation experiments are not included in the main paper. The Supplementary Material offers additional information that may interest readers, including different hyper-parameter configurations and the robustness of our method with different random seeds.

\begin{table}
\centering
\caption{Single-image evaluation results. We use five seeds and then take the mean and standard deviation. The best performer (prioritize the mean) is highlighted by \textbf{\textcolor{gred}{red}}.} 
\begin{tabular}{lccc}
\toprule
Method & SIQS-Q & SIQS-G & CCS (\%)          \\
\midrule
UDIS+R & 9.40{\tiny $\pm$ 0.08} & 9.27{\tiny $\pm$ 0.05} & 89.64{\tiny $\pm$ 0.96}   \\
UDISplus+R & 9.49{\tiny $\pm$ 0.05} & 9.33{\tiny $\pm$ 0.04} & 90.17{\tiny $\pm$ 0.88}   \\
SRStitcher & 9.28{\tiny $\pm$ 0.07}  & 8.44{\tiny $\pm$ 0.09} & 91.15{\tiny$\pm$ 0.52}  \\
LeftRefill & 9.23{\tiny $\pm$ 0.07} & 8.51{\tiny $\pm$ 0.13} & 83.98{\tiny $\pm$ 1.85}  \\
RDIStitcher (Ours) &\textbf{\textcolor{gred}{ 9.54{\tiny $\pm$ 0.07}}} & \textbf{\textcolor{gred}{9.39{\tiny $\pm$ 0.05}}} & \textbf{\textcolor{gred}{91.23{\tiny $\pm$ 0.79}}} \\
\bottomrule
\end{tabular}
\label{tab:quantitative}
\end{table}

\subsection{Qualitative Evaluation}
\begin{figure*}[t]
\centering
\includegraphics[width=0.96\linewidth]{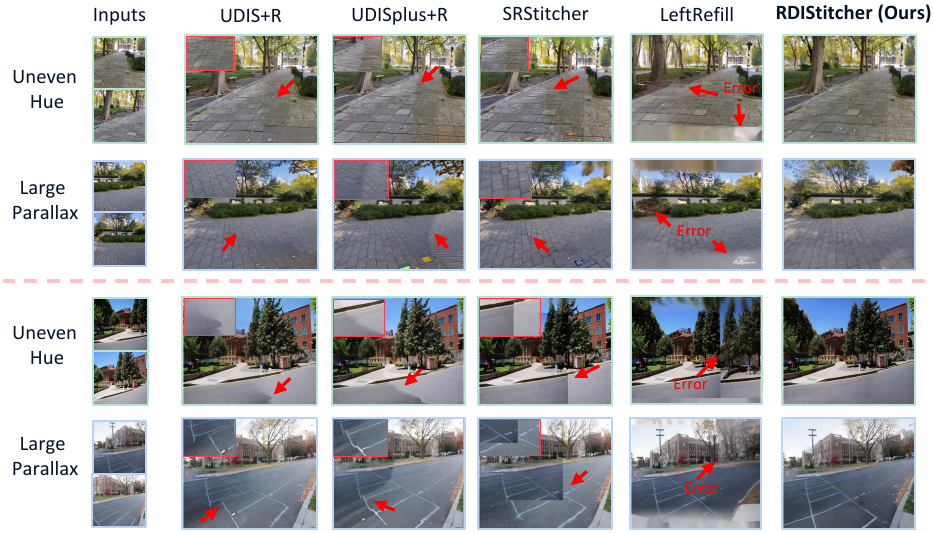}
\caption{Qualitative evaluation results. The upper half of the dotted line displays the results on UDIS-D, and the lower half is the results on traditional datasets. We highlight the areas with significant seams and errors using the local magnification and arrows. Special emphasis is placed on the error regions for the LeftRefill method, providing a thorough analysis of its performance limitations. Notice that the last example represents a hybrid challenge with uneven hue, large parallax, and zero-shot conditions. In this highly complex scene, our method shows exceptional performance, markedly surpassing that of previous methods. More results can be found in the supplementary Material.}
\label{fig:Qualitativeevaluation}
\end{figure*}

The qualitative evaluation results are shown in \textcolor{red}{Fig.}\ref{fig:Qualitativeevaluation}, highlighting the significant advantages of our method across various challenging scenarios. Notably, in the zero-shot experiments, our method demonstrates more obvious advantages, which verifies the generalization capability hypothesis proposed in \textcolor{red}{Sec.}\ref{sec:STM}.

\subsection{Consistency Evaluation}
To assess whether our method introduces content that deviates from the original images, we use Peak Signal-to-Noise Ratio (PSNR)$\uparrow$ and Structural Similarity Index (SSIM)$\uparrow$ as metrics to compare the consistency of the stitched images' fusion regions between our method and UDIS. Additionally, the comparison results between UDIS and UDIS++ are provided as references, as shown in the \textcolor{red}{Table.}\ref{tab:consistency}. The results demonstrate that our method achieves higher consistency in small parallax scenes $D_S$, large parallax scenes $D_L$, and the entire UDIS-D test set $D_F$, confirming its effectiveness in preserving content consistency. Detailed information on $D_S$ and $D_L$ and further discussion can be found in the Supplementary Material.

\begin{table}

\centering
\caption{Consistency evaluation results. We use five seeds and take the mean. The best performer is highlighted by \textbf{\textcolor{gred}{red}}.} 
\begin{tabular}{lcccc}
\toprule
\multicolumn{2}{l}{Method}                                  & $D_S$ & $D_L$ & $D_F$ \\
\midrule
\multirow{2}{*}{\begin{tabular}[c]{@{}l@{}}UDIS++ vs. UDIS\end{tabular}} & PSNR &  32.49  &  30.71  &   31.02 \\
                                                                             & SSIM &  0.747  &  0.531  &  0.602  \\
\multirow{2}{*}{\begin{tabular}[c]{@{}l@{}}Ours vs. UDIS\end{tabular}}   & PSNR &  \textbf{\textcolor{gred}{32.56}} & \textbf{\textcolor{gred}{31.42}}  & \textbf{\textcolor{gred}{31.66} } \\
                                                                             & SSIM &  \textbf{\textcolor{gred}{0.824}}  &  \textbf{\textcolor{gred}{0.736}}  &  \textbf{\textcolor{gred}{0.761}} \\
\bottomrule
\end{tabular}
\label{tab:consistency}
\end{table}

\subsection{MLLMs-based Metric Accuracy Evaluation}

Referring to work \cite{wu2024comprehensive}, we introduce  Spearman’s rank correlation coefficient (SRCC)$\uparrow$ and  Pearson linear correlation coefficient (PLCC)$\uparrow$ to evaluate the accuracy of the MLLMs-based metrics and compare them with the advanced NR-IQA methods including Q-align \cite{qalign}, Topiq \cite{topiq} and UNIQUE (UNIQ) \cite{zhang2021uncertainty} on a specially crafted stitched image quality dataset $D_{quality}$. The results are shown in the \textcolor{red}{Table.}\ref{tab:metricevaluation},which indicate that our proposed metrics are more in line with human visual perception. For more details about this experiment and $D_{quality}$, please refer to the Supplementary Material.

\begin{table}
\caption{MLLMs-based metric accuracy evaluation results. The best performer is highlighted by \textbf{\textcolor{gred}{red}}.} 
\begin{tabular}{lccccc}
\toprule
Metric & SIQS-Q & SIQS-G & Q-align & Topiq & UNIQ \\
\midrule
SRCC   &   \textbf{\textcolor{gred}{0.728}}     &     0.634   &     0.455    &    0.342   & 0.299 \\
PLCC   &     \textbf{\textcolor{gred}{0.685}}   &   0.525     &     0.412    &  0.335    & 0.242 \\
\bottomrule
\end{tabular}
\label{tab:metricevaluation}
\end{table}



\section{Conclusion}

In this paper, we propose an image stitching model based on reference-driven inpainting, which achieves remarkable seamless stitching results on challenging scenes. In addition, we propose the MLLMs-based metrics for stitched images. Extensive experiments demonstrate the effectiveness of our method and the proposed metrics, and our method highlights its strong generalization performance in the zero-shot scenario.




{
    \small
    \bibliographystyle{ieeenat_fullname}
    \bibliography{main}
}

\clearpage
\setcounter{page}{1}
\maketitlesupplementary
\appendix

\section{Overview}
\label{sec:supOverview}
In the supplementary material, we provide the following contents:
\begin{itemize}
\item Source code for our method, including train, inference, and MLLMs-based metrics. Please see the \textcolor{blue}{\textit{code}} document;
\item Survey details of \textcolor{red}{Fig.}\ref{fig:suvey} (\textcolor{red}{Sec.}\ref{sec:supSurvey});
\item More Details of RDIStitcher-R (\textcolor{red}{Sec.}\ref{sec:RDIStitcher-R}); 
\item More Information of Consistency Evaluations (\textcolor{red}{Sec.}\ref{sec:Consistency}); 
\item More Information of Ablation Studies (\textcolor{red}{Sec.}\ref{sec:ablations}); 
\item More Details of MLLMs-based Metrics (\textcolor{red}{Sec.}\ref{sec:MLLMSmetrics});
\item More Information of Metric Evaluation (\textcolor{red}{Sec.}\ref{sec:Metricevaluation});
\item More Results of Qualitative Evaluations (\textcolor{red}{Sec.}\ref{sec:mqevaluations}).
\end{itemize}

\section{Survey Details} \label{sec:supSurvey}

We first select 50 challenging samples exhibiting uneven hue, large parallax, and mixed scenarios (containing both challenges) from the UDIS-D \cite{nie2021unsupervised}, APAPdataset \cite{zaragoza2013projective}, REWdataset \cite{li2017parallax}, and SPWdataset \cite{liao2019single}. After applying the UDIS \cite{nie2021unsupervised}, UDIS++ \cite{nie2023parallax}, and SRStitcher \cite{xie2024reconstructing} to fuse these samples, we task four volunteer participants with rating the visual quality of the stitched images. Specifically, Volunteers provided scores of either \textit{good} or \textit{bad} based on two criteria: brightness and hue smoothness for uneven hues and content continuity and artifact severity in seam regions for large parallax. According to the evaluation results, methods achieving over 75\% \textit{good} ratings for a challenge receive a red smiley face, 50\%-75\% receive a yellow smiley face, 25\%-50\% a green face and less than 25\% a blue sad face.

\section{More details of RDIStitcher-R} \label{sec:RDIStitcher-R}

\subsection{Implementation of RDIStitcher-R}

RDIStitcher-R is a variant of RDIStitcher that only performs rectangling tasks, so its implementation is much simpler. 

Similar to \textcolor{red}{Sec.}\ref{sec:STM}, we get the pre-knowledge mask distribution $\mathbf{M} =\{(M^1_{wr},M^1_{wt}),(M^2_{wr},M^2_{wt}),...,(M^N_{wr},M^N_{wt})\}$, where $N$ is the number of samples in UDIS-D training set $D_{train}$. $I_{sg}$ is a single-view image in the $D_{train}$ and $(M^i_{wr},M^i_{wt}), i \in [1,N]$ is a random mask set from $\mathbf{M}$, the pseudo-fusion image $\tilde{I}_{pf}$ can be obtained by \textcolor{red}{Eq.}\ref{eq:pseudosufsion}.

\begin{equation}
\tilde{I}_{pf}  = I_{sg} \odot (M^i_{wr} \vee M^i_{wt}).
\label{eq:pseudosufsion}
\end{equation}

The coarse rectangling fusion image can be obtained by \textcolor{red}{Eq.}\ref{eq:crpseudosufsion}.

\begin{equation}
\tilde{I}^{cr}_{pf} = \mathsf{Telea}(\tilde{I}_{pf}, M^i_{wr}\vee M^i_{wt},R).
\label{eq:crpseudosufsion}
\end{equation}

We also produce the gradient mask by \textcolor{red}{Eq.}\ref{eq:gmpseudosufsion}.
\begin{equation}
\mathfrak{M} ^{i}_{gm} = \mathsf{Blur}(\mathsf{Dilation}(M^i_{wr} \vee M^i_{wt},K_d),K_g).
\label{eq:gmpseudosufsion}
\end{equation}

\begin{figure}[ht]
\centering
\includegraphics[width=0.96\linewidth]{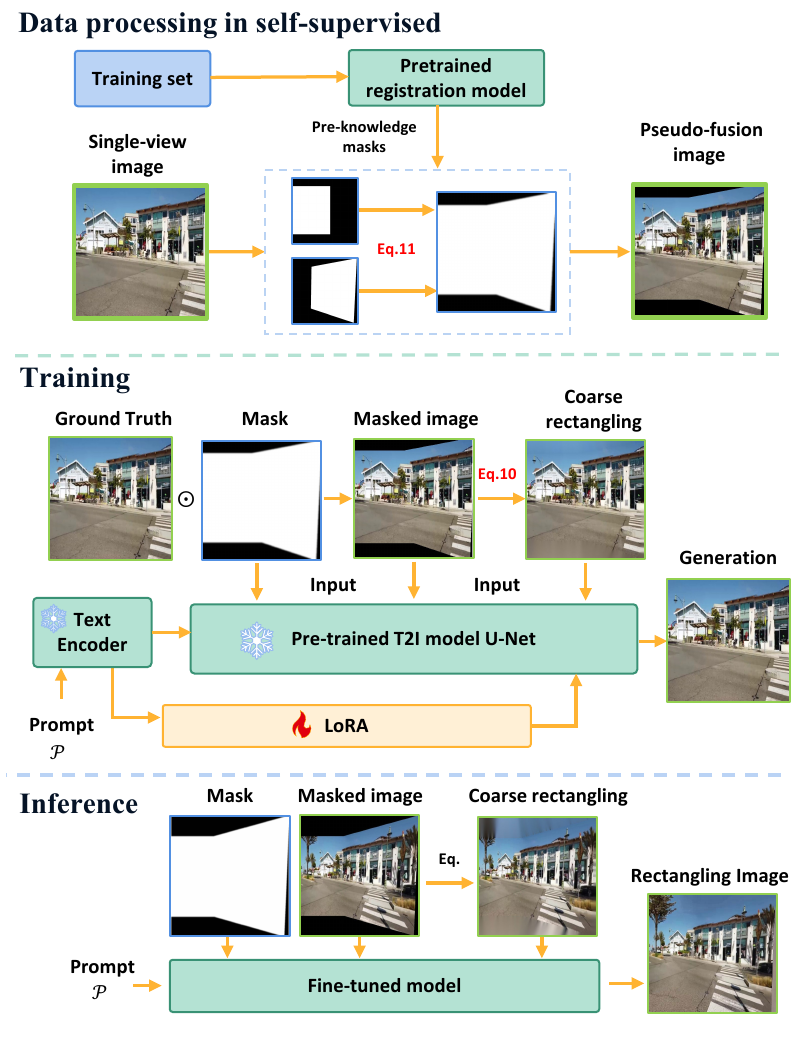}
\caption{The framework of RDIStitcher-R.}
\label{fig:RDIStitcherR}
\end{figure}

An overview of the architecture of RDIStitcher-R is shown in \textcolor{red}{Fig.}\ref{fig:RDIStitcherR}. 

\subsection{Ablation Study of RDIStitcher-R}

RDIStitcher-R operates by fine-tuning the pre-trained T2I model to establish the relationship between the coarse rectangling image and the detailed image. 

\textcolor{red}{Table.}\ref{tab:asRDIStitcherR} shows the ablations of USID+RDIStitcher-R (UDIS+R) based on the content consistency score CCS \cite{xie2024reconstructing} on the UDIS-D test set. The comparisons include UDIS+stable diffusion inpainting model \cite{stablediffusion2inpainting} without fine-tuning (UDIS+SD2) and UDIS+RDIStitcher-R without coarse rectangling (UDIS+RWCR). The results demonstrate that RDIStitcher-R effectively reduces the likelihood of abnormal content.

\begin{table}
\centering
\caption{Ablation study of RDIStitcher-R. The best performer is highlighted by \textbf{\textcolor{gred}{red}}.} 
\begin{tabular}{lccc}
\toprule
Method & UDIS+SD2 & UDIS+RWCR & UDIS+R         \\
\midrule
 CCS (\%) & 85.97{\scriptsize $\pm$ 1.33} & 87.68{\scriptsize $\pm$ 1.17} & \textbf{\textcolor{gred}{89.64{\tiny $\pm$ 0.96}}}   \\
\bottomrule
\end{tabular}
\label{tab:asRDIStitcherR}
\end{table}

\subsection{Performance of RDIStitcher-R}
RDIStitcher-R achieves an SSIM score of 0.763 and a PSNR of 23.56 on the DIR-D dataset, indicating that its performance is situated between the DeepRectangling \cite{nie2022deep} and Recdiffusion \cite{zhou2024recdiffusion}. However, RDIStitcher-R holds a significant advantage in its capability to be trained on unlabeled datasets, making it more versatile in scenarios where labeled data are scarce or difficult to obtain. Additionally, its hardware requirements are notably lower compared to those of Recdiffusion. The performance gap between RDIStitcher-R and Recdiffusion is acceptable in view of these advantages.

\section{More Information of Consistency Evaluations} \label{sec:Consistency}

\subsection{Information of $D_S$ and $D_L$}
To check the consistency, we manually made the small parallax scenes dataset $D_S$ and the large parallax scenes dataset $D_L$, both of which are sub-datasets contains 100 samples of the UDIS-D \cite{nie2021unsupervised}, whose details are shown in \textcolor{blue}{\textit{datasets/ConsistencyDataset.csv}} in Supplementary Material, where the names of the sample are consistent with those in UDIS-D.

\subsection{Experimental results based on masked SSIM and masked PSNR}

In addition to the consistency experiments in the main paper, we observed that the parallax regions of the UDIS results are very prone to artifacts leading to low confidence in this region. Therefore, we conduct further tests focusing on the consistency of regions excluding the parallax regions. Specifically, we use the masked Structural Similarity Index (mSSIM) and masked Peak Signal-to-Noise Ratio (mPSNR) as metrics to evaluate image coherence after masking out the parallax regions. The mSSIM and mPSNR were calculated as follows:

\begin{align}
mSSIM & = SSIM(I_{s}^{1}\odot M_{o},I_{s}^{2}\odot M_{o}),\\
mPSNR & = PSNR(I_{s}^{1}\odot M_{o},I_{s}^{2}\odot M_{o}),
\end{align}

where, $I_s^1$ and $I_s^2$ are two images stitched by different methods, such as UDIS and our method. And, $M_{o}$ is the overlap mask, that is $M_{o}=1-(M_{wr}\& M_{wt})$, where $\&$ is bitwise and operation.

The results of the consistency experiments based on mSSIM and mPSNR are shown in the \textcolor{red}{Table.}\ref{tab:mSSIMmPSNR}. Note that we don not test the results on $D_S$, because in the scenarios with small parallax, the masked area of $M_o$ becomes significantly large, leading to most areas of the image without content, thus losing the significance of comparison experiments on the $D_S$ dataset. 

\begin{table}
\centering
\caption{Consistency evaluation results by mSSIM and mPSNR. We use five seeds and take the mean. The best performer is highlighted by \textbf{\textcolor{gred}{red}}.} 
\begin{tabular}{lccc}
\toprule
\multicolumn{2}{l}{Method}                                  &  $D_L$ & $D_F$ \\
\midrule
\multirow{2}{*}{\begin{tabular}[c]{@{}l@{}}UDIS++ vs. UDIS\end{tabular}} & mPSNR &    32.90  &   35.91 \\
                                                                             & mSSIM &   0.701  &  0.838  \\
\multirow{2}{*}{\begin{tabular}[c]{@{}l@{}}Ours vs. UDIS\end{tabular}}   & mPSNR &   \textbf{\textcolor{gred}{33.19}}  & \textbf{\textcolor{gred}{36.04} } \\
                                                                             & mSSIM &    \textbf{\textcolor{gred}{0.812}}  &  \textbf{\textcolor{gred}{0.881}} \\
\bottomrule
\end{tabular}
\label{tab:mSSIMmPSNR}
\end{table}

\section{More Information of Ablation Studies} \label{sec:ablations}

\subsection{LoRA Configurations} 

LoRA \cite{hu2022lora} is considered one of the most influential technologies in recent years, with numerous studies exploring its mechanisms. Due to the limited space of the main paper, all parameters of LoRA are not introduced in detail. Here, we provide additional parameters: rank 8, alpha constant 16, dropout rate 0.1, and bias none. For more details, please refer to the source code we have provided.

We conducted an ablation test by varying the LoRA rank while keeping other parameters fixed. The results are shown in the \textcolor{red}{Table.}\ref{tab:lorarankabla}. However, we quickly realized that this isolated exploration does not definitively demonstrate that a LoRA rank 8 represents the optimal parameter setting for our specific task. A comprehensive exploration of LoRA cnfiguration is complex and requires significant hardware resources, which is beyond the scope of this study. Therefore, we provide a reference set of LoRA parameter values in the main paper, but have not pursued any further ablation analysis regarding the LoRA settings.

\begin{table}[h]
\centering
\caption{Ablation study of LoRA rank. The best performer is highlighted by \textbf{\textcolor{gred}{red}}.} 
\begin{tabular}{lccc}
\toprule
Method & Rank 4 & Rank 8 & Rank 16         \\
\midrule
 CCS (\%) & 90.54{\scriptsize $\pm$ 0.92} & \textbf{\textcolor{gred}{91.23{\tiny $\pm$ 0.79}}} & 89.35{\scriptsize $\pm$ 0.83}   \\
\bottomrule
\end{tabular}
\label{tab:lorarankabla}
\end{table}

\subsection{Seed robustness}
\begin{figure*}[h]
\centering
\includegraphics[width=1.0\linewidth]{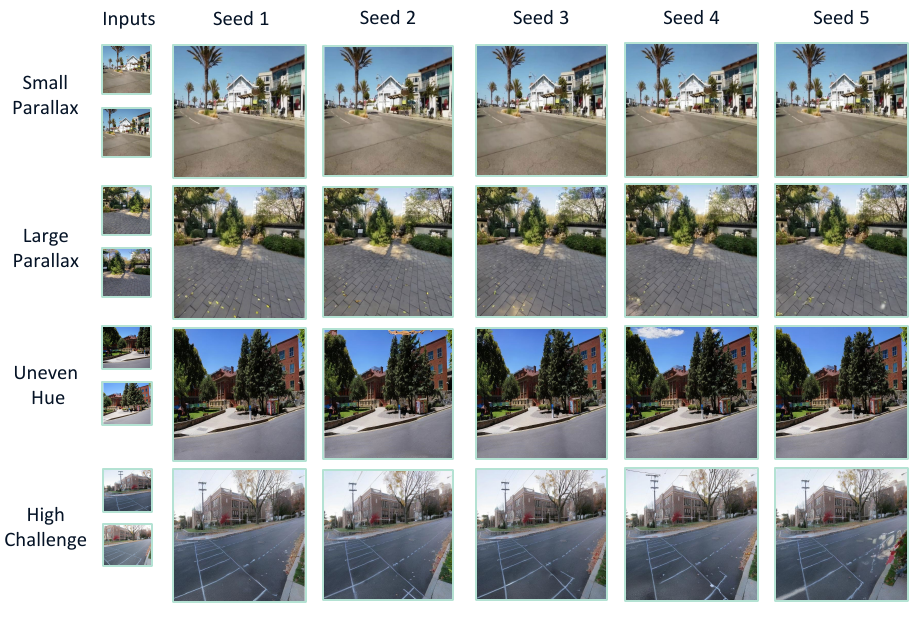}
\caption{Qualitative results on the different seeds.}
\label{fig:seedresults}
\end{figure*}

 We present the qualitative evaluation results with different random seeds in \textcolor{red}{Fig.}\ref{fig:seedresults}. In cases of small parallax, the stitched image produced by our method maintains high consistency. For large parallax, differences in the fusion region are insignificant, and slight changes occur in the rectangling region. Importantly, our method does not introduce significant abnormal content, unlike the original T2I model. In zero-shot tests, the stitched images of uneven hue and small parallax scenes also maintain good consistency, with only minor changes in the rectangling region. In high challenge scenarios, while content changes in the stitched images are more pronounced, our method shows clear advantages in brightness smoothing and content connection compared to previous methods (see \textcolor{red}{Fig.}\ref{fig:Qualitativeevaluation} of main paper for comparison). 

It is worth noting that the inherent characteristics of the generation-based model can lead to occasional instability, resulting in some failure cases. However, our experiments indicate that the failure rate of our method is very low and within an acceptable range. Future work will focus on further enhancing reliability .

\section{More Details of MLLMs-based Metrics} \label{sec:MLLMSmetrics}

\subsection{Full Example of MICQS Evaluation}

Due to space limitations, the evaluation prompt and reasons of MICQS evaluation in the main paper is truncated. Here, \textcolor{red}{Table.}\ref{tab:fullmicqs} gives a complete example of MICQS evaluation based on Qwen-VL-Max \cite{Qwen-VL}.

\renewcommand\arraystretch{1.0}
\begin{table}[h]
\centering
\caption{A full example of MICQS evaluated by Qwen-VL-Max.}
\label{tab:fullmicqs}
\begin{tabular}{p{3in}}
\toprule
\multicolumn{1}{c}{
    \includegraphics[width=0.25\linewidth]{figs/SIQStable.jpg}
    \hspace{25pt}
    \includegraphics[width=0.25\linewidth]{figs/MICQStable.jpg}
} \\
\midrule
\textbf{Evaluation Prompt} \\
I need you to become a stitched image quality assessment evaluator. Compare the input two stitched images, includes seam, brightness transition, distortion, clear and abnormal content. Choose which one you think is better. There are two choices: image 1 (left) is better, or image 2 (right) is better.: \\
\midrule
\textbf{Reason} \\
- \textbf{Seam} Image 2 has a smoother transition between the stitched sections compared to image 1, where the seam is more noticeable. \\
- \textbf{Brightness transition} Image 2 has a more consistent brightness level across the entire image, while image 1 shows a noticeable difference in brightness between the sections. \\
- \textbf{Distortion} There are no noticeable distortions in the both images. \\
- \textbf{Clear} Image 2 appears clearer overall. \\
- \textbf{Abnormal content} Image 2 has less noticeable artifacts or abnormalities in the content. \\
\midrule
\textbf{Conclusion} \\
\textbf{\textcolor{gblue}{Image 2 (right)}} is better.\\
\bottomrule
\end{tabular}
\end{table}

\subsection{Full Example of SIQS Evaluation}

Due to space limitations, the evaluation prompt of SIQS evaluation in the main paper is truncated. Here, \textcolor{red}{Table.}\ref{tab:fullsiqs} gives a complete example of SIQS evaluation based on Qwen-VL-Max \cite{Qwen-VL}.

\subsection{Experiment Settings of MLLMs-based Metrics}

To evaluate the stitched images quality, we use the API provided by the MLLMs service. Occasionally, errors occurred for unknown reasons led to the absence of correct result returns. Therefore, for each test stitched image, we submit multiple test requests until a valid return value is obtained for evaluation.

\subsection{The cost of MLLMs-based Metrics}

All MLLMs used in this paper offer free access through their APIs, with operating costs below \$0.02 per thousand tokens for both input and output. For comparison, if we were to manually evaluate the UDIS-D test set (1,106 samples), it would take as fast as an hour and a half to perform an evaluation assuming that a stitched images is evaluated for 5 seconds (the actual evaluation time would be longer because the human eye needs a rest period). Therefore, the use of MLLM-based metrics proves to be not only more economical, but also significantly faster than manual inspection.  

\subsection{Limitations of MLLMs-based Metrics}

\textbf{Prompt sensitivity} Different MLLMs exhibit varying sensitivities to prompts. As depicted in \textcolor{red}{Table.}\ref{tab:quantitative}, GLM-4V-Plus \cite{glm2024chatglm} scores significantly lower than other methods for SRStitcher \cite{xie2024reconstructing} and LeftRefill \cite{cao2024leftrefill}, as blurred is a key scoring criterion. Although SRStitcher and LeftRefill may stitch images with numerous local blur instances, Qwen-VL-Max \cite{Qwen-VL} is relatively insensitive, not displaying a significant score reduction. Therefore, the construction of prompts can significantly impact the evaluation performance of the MLLMs.

\noindent
\textbf{Concept limitation} In the early design, we intended to include artifacts in the evaluation criteria of MLLMs-based metrics. However, incorporating \textit{Artifact}-related prompt texts led to numerous false evaluations. Through the analysis of the results, we found that the MLLMs used in our paper actually does not know the concept of \textit{Artifact}, resulting in significant misjudgments. This substantial error prompted us to replace the \textit{Artifact}-related prompt texts with \textit{Abnormal Content}. Although test results indicate that MLLMs struggle to assess what constitutes abnormal content, they at least avoid major misjudgments. It is evident that the current knowledge domain of MLLMs we utilize in the paper has considerable limitations in evaluating the quality of stitched images.

\section{More Information of Metric Evaluation} \label{sec:Metricevaluation}

\subsection{Information of $D_{quality}$}

We hand-crafted a stitched image dataset for measuring image quality, named $D_{quality}$, containing 50 different stitched images, whose details can be found in \textcolor{blue}{\textit{datasets/QualityDataset.zip}}. To enrich the diversity of $D_{quality}$, we intentionally incorporated examples where various stitching methods introduced errors, including severe artifacts, distortions, structural deformations, content anomalies, and noticeable seams. In addition, the dataset includes a selection of well-stitched images to evaluate the ability of the metric to accurately identify high quality stitching results.

\subsection{More Details of MLLMs-based Metric Accuracy Evaluation}
Since the image stitching datasets are unlabeled, traditional reference-based metrics for assessing image quality cannot be applied. To address this challenge, we adopt a methodology inspired by other image quality evaluation studies, establishing an evaluation benchmark based on mean user scores. Specifically, we invite four volunteers to score the samples in $D_{quality}$ according to the criteria showed in \textcolor{red}{Table.}\ref{tab:fullsiqs}. Then, we calculate the SRCC and PLCC between the mean user scores and comparison metrics. These correlation coefficients serve as a measure to evaluate the accuracy and reliability of each metric in reflecting the stitched image quality, ensuring its alignment with human perceptual judgement.

\subsection{Visualized Analysis}

\textcolor{red}{Fig.}\ref{fig:supfigmorequalia1} presents a visual comparison among metrics. We show the most error-prone cases for three types of image stitching, where images with green borders represent the correctly stitched images. Conversely, images with blue borders indicate instances where the stitching process resulted in errors or suboptimal output.

The first example is the seam in the stitched image caused by uneven hue. It is evident that Q-Align, Topiq, and UNIQ cannot accurately identify the image with the superior stitching quality in this scenario. According to the design logic of the general NR-IQA models, they prefer the brighter images, leading to the judgement error. While, in our MLLMs-based metrics, prompt can be used to guide the assessment criteria. By incorporating seam-related prompts, our metrics can correctly identify the higher-quality stitched image, thereby overcoming the limitations faced by current NR-IQA models. 

The second example is the presence of Artifacts, all evaluated metrics can correctly identify the better stitched image. The appearance of artifacts closely resembles that of ghosting effects, leading to content blurring, which is an important factor in general NR-IQA research. Therefore, NR-IQA method can handle such cases well, and our proposed metrics can also handle such scenarios well. 

The third example is a misalignment scene, where it can be seen that the wires are not connected correctly. Unfortunately, all metrics fail to correctly judge better stitched images in this scenario, and UNIQ shows the most significant performance degradation. Despite the broken wires being readily apparent to human observers, it is clear that such an error highly related to image semantics is still a major challenge for existing image quality assessment models. It suggests that future research should explore the integration of more semantically-aware factors into these metrics to improve their efficacy in recognizing and evaluating such defects.

\begin{figure}[h]
\centering
\includegraphics[width=1.0\linewidth]{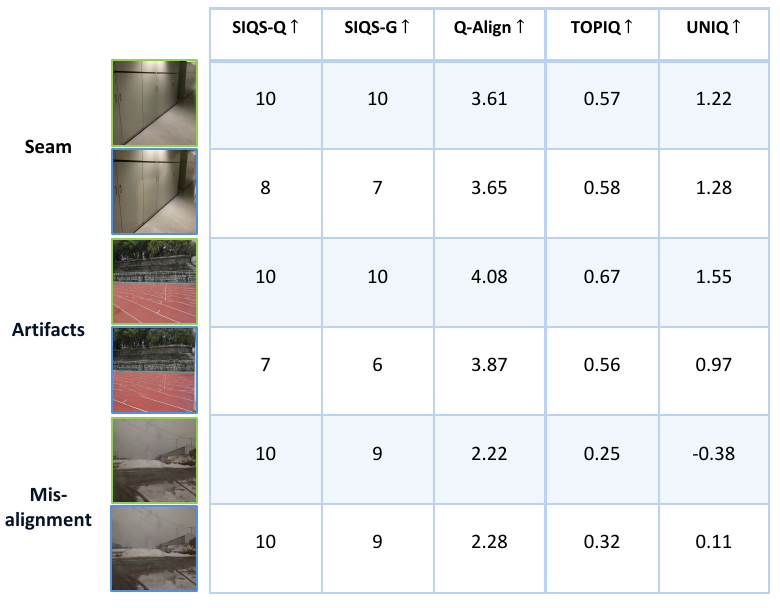}
\caption{Qualitative results on the different metrics.}
\label{fig:visiualmetric}
\end{figure}

\section{More Results of Qualitative Evaluations} \label{sec:mqevaluations}

Due to space limitations, we only present a limited number of qualitative evaluation results in the main paper. Here, to further examine the generalization capabilities of the RDIStitcher, We provide additional qualitative results obtained on the UDIS-D \cite{nie2021unsupervised}, APAPdataset \cite{zaragoza2013projective}, REWdataset \cite{li2017parallax}, and SPWdataset \cite{liao2019single}.

\begin{figure*}[ht]
\centering
\includegraphics[width=0.96\linewidth]{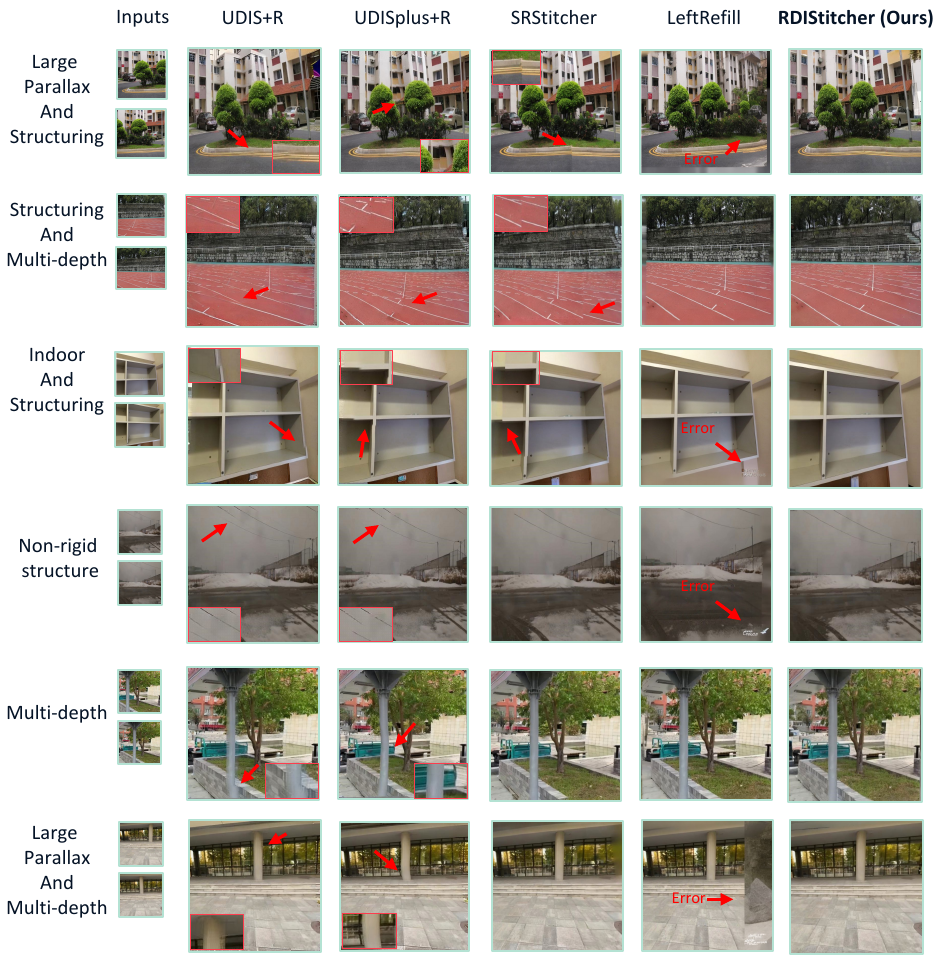}
\caption{Additional qualitative results for the challenging scenarios.}
\label{fig:supfigmorequalia1}
\end{figure*}

\textcolor{red}{Fig.}\ref{fig:supfigmorequalia1} illustrates the qualitative results for more challenging scenarios. The first example is the large parallax and structuring scene, where the difficulty is how to maintain the original structure during stitching. UDISplus+R shows obvious structural distortion when dealing with such scenes, while SRStitcher shows obvious seams. In contrast, our method achieves a better smoothing effect in the seam region while preserving the object structure; The second example is the structuring and multi-depth layer scene, where the challenge is aligning different depth layers uniformly. Here, it is evident that only the background is correctly aligned, while the foreground runway exhibits misalignment, leading to severe artifacts in UDIS+R. Similarly, both UDISplus+R and SRStitcher show clear misalignments in the white lines of the runway. In contrast, our method excels in this scenario by maintaining the structural integrity of the runway lines; The fourth scene contains some non-rigid transformation objects, which are wires in the sky. These wires do not adhere to rigid transformation principles during the imaging process. As a result, commonly used registration methods, such as homography, face significant challenges in accurately aligning these elements, leading to the problem of abnormal number of wires in UDIS+R and UDISplus+R. In contrast, our method handles this scenario well, demonstrating the robustness of our scheme in dealing with registration errors; The fifth example is also a multi-depth layer scene, UDIS+R struggles with significant artifacts, while UDISplus+R causes severe distortions, particularly affecting the pillars' structure. In contrast, inpainting-based methods handle this scene effectively, demonstrating their advantage in maintaining structural integrity and reducing distortion. This highlights the superiority of inpainting-based methods in this kind of scene; The sixth example is the scene of large parallax and multi-depth layers, in which UDIS+R still exhibits artifacts, while UDISplus+R is distorted, resulting in structural distortion of the pillars. In contrast, our method demonstrates sustained high performance in such challenging scenarios, effectively maintaining structural accuracy and minimizing distortions. 

\begin{figure*}[ht]
\centering
\includegraphics[width=0.96\linewidth]{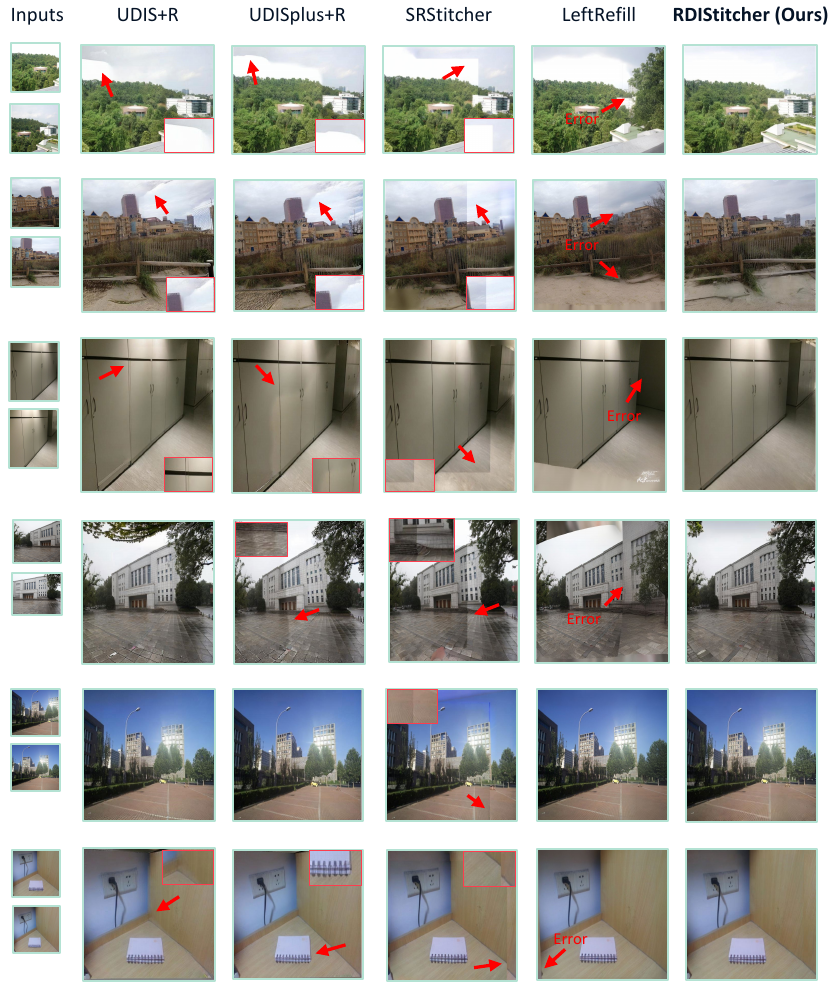}
\caption{Additional qualitative results on APAPdataset \cite{zaragoza2013projective}, REWdataset \cite{li2017parallax}, and SPWdataset \cite{liao2019single}.}
\label{fig:supfigmorequalia2}
\end{figure*}

\textcolor{red}{Fig.}\ref{fig:supfigmorequalia2} presents the results under more datasets, including APAPdataset \cite{zaragoza2013projective}, REWdataset \cite{li2017parallax}, and SPWdataset \cite{liao2019single}.

\renewcommand\arraystretch{1.0}
\begin{table*}[h]
\centering
\caption{A full example of SIQS evaluated by Qwen-VL-Max.}
\label{tab:fullsiqs}
\begin{tabular}{p{6in}}
\toprule
\multicolumn{1}{c}{
    \includegraphics[width=0.25\linewidth]{figs/SIQStable.jpg}
} \\
\midrule
\textbf{Evaluation Prompt} \\

I need you to become a stitched image quality assessment evaluator. The evaluation process should be as objective and impartial as possible, giving specific ratings and reasons, including seam, brightness transition, distortion, clear and abnormal content, each aspect 2 points. \\
\textbf{1. Whether there are seams in the image (2 points).} \\
\textit{score 2: the image is smooth without obvious boundaries or misalignment;} \\
\textit{score 1: there are slightly visible boundaries in the image, but overall look well;}  \\
\textit{score 0: there are obvious borders or dislocations in the image, affecting the overall look and feel.} \\
\textbf{2. Whether there are brightness transitions in the image (2 points).} \\
\textit{score 2: the brightness transition of image is smooth;} \\
\textit{score 1: the light and shade changes in the image are a bit unnatural;}  \\
\textit{score 0: the light and shade changes in the image are very abrupt.} \\
\textbf{3. Whether there are distortions in the image (2 points).} \\
\textit{score 2: no distortion in the image;} \\
\textit{score 1: there are a few structural anomalies of straight lines in the image;}  \\
\textit{score 0: there are noticeably distortions, such as distorted pillar, brick, and building construction.} \\
\textbf{4. Whether the image is clear and blurred (2 points).}  \\
\textit{score 2: the image is clear, the details are visible, and there is no blur;} \\
\textit{score 1: the resolution of the image is good, but slightly blurred;}  \\
\textit{score 0: the image is blurred and the details are not clear.} \\
\textbf{5. Whether the image is natural (2 points).} \\
\textit{score 2: the image is natural with out abnormal content; } \\
\textit{score 1: there are some places in the image that is not in harmony with the main content;}  \\
\textit{score 0: There are a lot of abnormal content in the image such as strange texture and non-semantic image.} \\
\midrule
\textbf{Reason} \\
- \textbf{Seam} The image shows slightly visible boundaries, particularly noticeable in the sky and along the edges of the buildings. (\textbf{\textcolor{gred}{1 points}}). \\
- \textbf{Brightness transition} The brightness transition in the image is relatively smooth. (\textbf{\textcolor{gred}{2 points}}). \\
- \textbf{Distortion} There are no noticeable distortions in the image. The straight lines of the buildings and the street appear to be accurate. (\textbf{\textcolor{gred}{2 points}}). \\
- \textbf{Clear} The buildings, palm trees, and streets are all distinguishable, and the details are visible. (\textbf{\textcolor{gred}{2 points}}). \\
- \textbf{Abnormal Content} The image contains some unnatural elements, such as the seams. (\textbf{\textcolor{gred}{1 points}}). \\
\midrule
\textbf{Score} \\
The overall impression is that the image is a stitched panorama with noticeable flaws. (\textbf{\textcolor{gblue}{8 points}}).\\
\bottomrule
\end{tabular}
\end{table*}

\end{document}